\UseRawInputEncoding
\pdfoutput=1
\documentclass{IEEEtran}
\usepackage{cite}
\usepackage{amsmath,amssymb,amsfonts}
\usepackage{algorithmic}
\UseRawInputEncoding
\usepackage{graphicx}
\usepackage{textcomp}
\usepackage{multirow}
\def\BibTeX{{\rm B\kern-.05em{\sc i\kern-.025em b}\kern-.08em
		T\kern-.1667em\lower.7ex\hbox{E}\kern-.125emX}}
\usepackage{balance}
\usepackage{setspace} 
\usepackage{cite}
\usepackage{amsmath,amssymb,amsfonts}
\usepackage{algorithmic}
\usepackage{caption}   
\usepackage{subfigure} 

\usepackage{upgreek}
\usepackage{bm}
\usepackage{url}

\begin{document}
	\title{QS-ADN: Quasi-Supervised Artifact Disentanglement Network for Low-Dose CT Image Denoising by Local Similarity Among Unpaired Data}
	\author{Yuhui Ruan, Qiao Yuan, Chuang Niu, Chen Li, Yudong Yao, \IEEEmembership{Fellow, IEEE},  Ge Wang, \IEEEmembership{Fellow, IEEE}, and Yueyang Teng*
		
		\thanks{This work was supported by the Natural Science Foundation of Liaoning Province (Grant No. 2022-MS-114). (Corresponding author: Yueyang Teng.)}
		\thanks{The first two authors contributed equally to this work.}
		\thanks{R. Ruan, Q. Yuan, and C. Li are with the College of Medicine and Biological Information Engineering, Northeastern University, Shenyang 110169, China.}
		\thanks{C. Niu is with the Department of Biomedical Engineering, Rensselaer Polytechnic Institute, Troy, NY 12180, USA.}
		\thanks{Y. Yao is with the Department of Electrical and Computer Engineering, Stevens Institute of Technology, Hoboken, NJ 07030 USA.}
		\thanks{G. Wang is with the Department of Biomedical Engineering, Rensselaer Polytechnic Institute, Troy, NY 12180, USA.}
		\thanks{Y. Teng is with the College of Medicine and Biological Information Engineering, Northeastern University, Shenyang 110169, China, and with the Key Laboratory of Intelligent Computing in Medical Image, Ministry of Education, Shenyang 110169, China. (e-mail: tengyy@bmie.neu.edu.cn).}}
	\maketitle
	\begin{abstract}
		Deep learning has been successfully applied to low-dose CT (LDCT) image denoising for reducing potential radiation risk. However, the widely reported supervised LDCT denoising networks require a training set of paired images, which is expensive to obtain and cannot be perfectly simulated. Unsupervised learning utilizes unpaired data and is highly desirable for LDCT denoising. As an example, an artifact disentanglement network (ADN) relies on unparied images and obviates the need for supervision but the results of artifact reduction are not as good as those through supervised learning.
		An important observation is that there is often hidden similarity among unpaired data that can be utilized. This paper introduces a new learning mode, called quasi-supervised learning, to empower the ADN for LDCT image denoising.
		For every LDCT image, the best matched image is first found from an unpaired normal-dose CT (NDCT) dataset. Then, the matched pairs and the corresponding matching degree as prior information are used to construct and train our ADN-type network for LDCT denoising.
		The proposed method is different from (but compatible with) supervised and semi-supervised learning modes and can be easily implemented by modifying existing networks. The experimental results show that the method is competitive with state-of-the-art methods in terms of noise suppression and contextual fidelity. The code and working dataset are publicly available at \url{https://github.com/ruanyuhui/ADN-QSDL.git}.
	\end{abstract}
	
	
	\section{Introduction}
	\label{sec:introduction}
	\IEEEPARstart{M}{edical} images are a prerequisite for many clinical diagnostic and therapeutic tasks. Common medical imaging modalities include X-ray radiography, computed tomography (CT), magnetic resonance imaging (MRI), nuclear and ultrasound imaging. Among them, CT has the advantages of high resolution, geometric accuracy, fast speed, and relatively low cost, which are often the first imaging study before an intervention is performed. 
	However, it cannot be ignored that X-rays produce ionizing radiation during a CT scan. When the X-ray radiation dose is absorbed by the human body, it may potentially induce abnormal metabolism or even genetic damage and cancer \cite{Ref1,Ref11}. The use of low-dose CT (LDCT) in practice can effectively reduce the radiation risk for patients but the resultant image noise and artifacts could compromise diagnosis \cite{Ref2}. Since the concept of LDCT was proposed \cite{Ref3}, a variety of methods were developed to suppress image noise and artifacts. These methods can be divided into three categories: projection domain filtering \cite{Ref4,Ref6,Ref7,Ref8}, iterative reconstruction \cite{Ref9,Ref10,12,Ref13,13} and image post-processing.
	
	The projection domain filtering methods process projections and then use a reconstruction method to produce a low-noise image. These methods include nonlinear filtering \cite{Ref4,Ref6} and statistical iterative methods \cite{Ref7,Ref8}. The main advantage of projection domain filtering is that it is easy to integrate the filter into existing CT systems. The iterative reconstruction methods use a likelihood function to associate the projections with a reconstructed image. 
	The key is to obtain suitable prior information, such as total variation (TV) \cite{Ref9}, nonlocal means \cite{Ref10}, dictionary learning \cite{12}, partial differentiation \cite{Ref13}, and low-rank matrix decomposition \cite{13}. 
	Such prior information can be incorporated into the objective function, and then the image is reconstructed in an iterative manner. These methods rely on projections \cite{Ref17}, which are generally inaccessible to most CT researchers unless they closely collaborate with a CT vendor. The image post-processing methods work on CT images only, without using projection data. Compared with the first two types of methods, the image post-processing methods can be applied after images are reconstructed, which are much more accessible than projection data. Image post-processing has been a hot topic in the field of LDCT image denoising. Along this direction, many excellent methods were developed, such as block-matching 3D filtering (BM3D) \cite{34} and dictionary learning-based filtering \cite{35}\cite{36}.
	
	Over recent years, deep learning has emerged as a new approach for image post-processing \cite{Ref25}. Many scholars  developed deep learning-based methods for estimating normal-dose CT (NDCT) images from LDCT images. These methods use different network structures, such as 2D convolutional neural networks (CNNs) \cite{Ref18}, 3D CNNs \cite{Ref19}, cascaded CNNs \cite{Ref20}, residual encoder-decoder CNNs \cite{Ref21}, and generative adversarial networks (GANs) \cite{Ref22}). They also used different loss functions to measure the similarity between outputs and labels, such as the mean square error, perceptual loss \cite{Ref23}, generative adversarial loss \cite{Ref24}, and Wasserstein distance. Most of these deep learning methods use the supervised learning mode \cite{Ref18,Ref21,Ref41,38}, which require paired LDCT and NDCT images with the corresponding pixels representing the same position in the same patient. However,
	paired real datasets require multiple scans, which results in not only an increased amount of manpower, material and financial resources but also an additional radiation dose. Under the principle of ALARA (as low as reasonably achievable) \cite{Ref11}, human studies with an excessive radiation dose are strongly discouraged. Furthermore, multiple scans of the same patient may be subject to motion artifacts and substantial registration errors. Hence, it is impractical to 
	obtain real paired datasets for supervised learning. In the AAPM Low-dose CT Challenge, low-dose CT images were simulated to compare different denoising algorithms. Although the detector-specific noise level can be realistically synthesized, scattering and other factors cannot be perfectly incorporated. In summary, it is an open issue how to construct a high-quality training dataset for low-dose CT image denoising.
	
	Many hospitals have accumulated large amounts of unpaired patient CT scans at different dose levels but in the supervised learning mode these images cannot be fully utilized due to their unpaired nature. As a result, unsupervised learning methods have attracted a major attention. GANs are among the most important unsupervised learning methods \cite{Ref22}. A GAN takes one image set as the learning target of another image set, greatly relaxing the requirement for image-level pairing.
	Park $et~al.$ \cite{Ref28} used a GAN and unpaired data to learn a generator that maps LDCT images to NDCT-like images.
	CycleGAN \cite{Ref32} as a variant of the generic GAN can realize image-to-image conversion between the input image domain and the target image domain with a cycle-wise consistency. Although it performs well for image conversion, some details may be distorted.
	As a further extension of CycleGAN, the artifact disentanglement network (ADN) is another well-known unsupervised learning method that maps unpaired low- and high-quality images into two latent spaces and then disentangles the contents and artifacts in the latent spaces, which supports image generation in different forms (artifact reduction, artifact transfer, self-reconstruction, etc.) \cite{Ref42}. For LDCT denoising, we need not only make image outputs look overall similar to NDCT images but also keep all the details as faithful as possible for diagnosis. However, the existing GAN-based methods are not satisfactory at preserving details.
	
\begin{figure}
		\centerline{\includegraphics[width=0.48\textwidth]{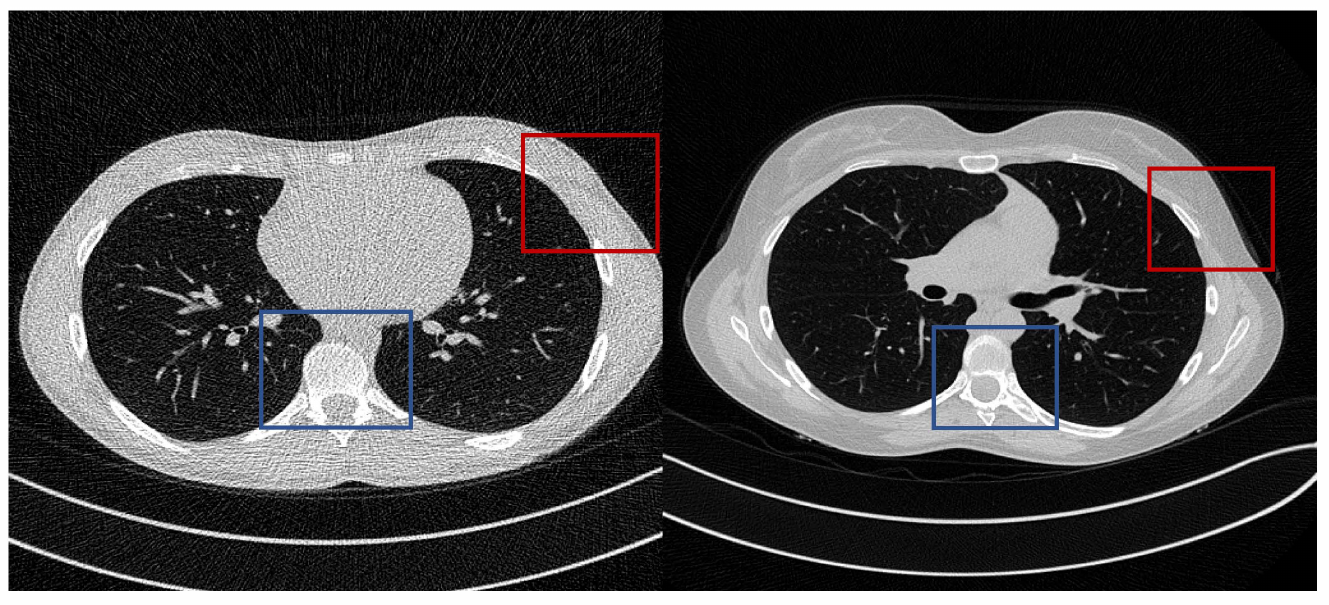}}
		\caption{Local similarity examples in two CT images from two patients under different imaging conditions respectively, where the boxes of the same color are anatomically consistent and structurally similar.}
		\label{fig1}
	\end{figure}

	Although the current GAN-based learning methods distinguish synthetic images from real images, some important information is still lost. In fact, even if two images come from two different patients, they still have some local similarity that can be leveraged to suppress image noise, based on the same idea behind non-local mean denoising methods.
	Fig. \ref{fig1} shows structural similarity between the sub-images boxed  in red and blue colors respectively, which is clearly higher than that between the red and blue regions.  Structural similarity is important prior information; for example, we applied it to achieve superresolution of PET images \cite{Ref36}, in which we proposed a new weakly supervised learning mode, referred to as quasi-supervised learning, for recovering high-resolution PET images from low-resolution counterparts by leveraging the similarity between unpaired data.
	
	This paper introduces the strategy of quasi-supervised learning into an ADN architecture for LDCT image denoising. The resulting network is called QS-ADN, which takes LDCT images as the input and unpaired NDCT images as the learning target. QS-ADN fully utilizes local similarity information of unpaired data to train the network, which is different from either supervised or unsupervised learning, and can be viewed as a new weakly supervised learning mode.
	It is not only applicable to unpaired data but also compatible with partially or fully paired data. In the next section, we present some background information. In the third section, we describe the proposed QS-ADN for LDCT denoising. In the fourth section, we report experimental results, which show the feasibility and effectiveness of the proposed method. In the last section, we discuss relevant issues and conclude the paper.
	
	\section{Background: Quasi-supervised Learning and ADN}
	This work introduces quasi-supervised learning into an ADN architecture for LDCT image denoising. In the following, let us review each of them.
	\subsection{Quasi-supervised learning}
	Our developed quasi-supervised learning method utilizes the hidden similarity between unpaired data as prior information for constructing a deep learning-based denoising network.
	The method consists of the following main steps:
	\begin{enumerate}
		\item A large number of unpaired LDCT and NDCT images are collected.
		\item They are divided into patches of meaningful structures.
		\item The best-matched pairs are identified from the LDCT and NDCT patches by their similarity.
		\item A network is trained on the matched patch pairs and their corresponding matching degrees.
	\end{enumerate}
	
	We take a general network as an example to explain the proposed quasi-supervised learning mode. Let $x$ and $y$ represent an LDCT and NDCT patch, respectively, where $x$ and $y$ can be paired or unpaired. Let $f(x,\theta)$ be a denoising operator implemented by a network with a vector of trainable parameters $\theta$, and $w(x, y)\in [0,~1]$ be the weight closely associated with the matching degree. Then, the quasi-supervised denoising objective can be expressed as
	\begin{equation}	\label{eq:qsl}
		\min \limits_\theta{\mathbb{E}_{(x,y)}}\left[ {w(x,y)\left\| {f(x,\theta ) -} \right.\left. y \right\|_2^2} \right]
	\end{equation}
	
	Note that quasi-supervised learning uses $w(x, y)$ to adjust the matching degrees of the patches. The larger the value of $w(x,y)$ is, the more similar the paired patches are. It can be viewed as a new weakly supervised learning mode. When $(x, y)$ are truly paired data, namely, $w(x, y)=1$, quasi-supervised learning becomes supervised learning. Therefore, it is compatible with supervised, unsupervised and semi-supervised learning strategies in the conventional sense.
	
	\subsection{ADN}
	As proposed by \cite{Ref42}, let $I^a$ be the domain of all artifact-affected CT images and $I$ be the domain of all artifact-free CT images. It is assumed that no paired dataset is available. We define a content (artifact-free) latent space $C$ and an artifact latent space $A$. Notably, the latent space $C$ is different from the observed space $I$.
	The ADN architecture contains an artifact-free image encoder $E_I$: $I$ $\rightarrow$ $C$ and a decoder $G_I$: $C$ $\rightarrow$ $I$
	and an artifact-affected image encoder $E_{I^a}$ = \{$E_{I^a}^c$ : $I^a$ $\rightarrow$ $C$; $E_{I^a}^a$ : $I^a$ $\rightarrow$ $A$\} and a decoder $G_{I^a}$ : $C$ $\times$ $A$ $\rightarrow$ $I^a$. The artifact-affected image encoder includes two subencoders which will be introduced below.
	The encoders map an image from the image domain to the latent space, and the decoders map a latent code back to the image domain.
	
	Specifically, given two unpaired images $x^a$ $\in$ $I^a$ and $y$ $\in$ $I$, $E_I$ and $E_{I^a}^c$ map the content components of $y$ and $x^a$ respectively to the content latent space $C$, and $E_{I^a}^a$ maps the artifact component of $x^a$ to the artifact latent space $A$. They can be formulated as follows:
	\begin{eqnarray}
		E_{I}(y)=c_y\in C,~
		E_{I^a}^c(x^a)=c_x \in C,~
		E_{I^a}^a(x^a)=a \in A
	\end{eqnarray}
	
	Then, the decoder $G_{I^a}$ takes a content code and an artifact code as its input and outputs an artifact-affected image. It is expected that decoding $c_x$ and $a$ should reconstruct $x^a$, and decoding $c_y$ and $a$ should add artifacts to $y$.
	\begin{equation}
		G_{I^a}(c_x,a)=\hat{x}^{a}\in I^a,~G_{I^a}(c_y,a)=\hat{y}^{a}\in I^a\\
	\end{equation}
	Similarly, the decoder $G_I$ takes a content code as its input and outputs an artifact-free image. Decoding $c_x$ should remove the artifacts from $x^a$, and decoding $c_y$ should reconstruct $y$.
	\begin{equation}
		G_I(c_x)=\hat{x}\in I,~ G_{I}(c_y)=\hat{y}\in I
	\end{equation}
	Note that $\hat{y}^{a}$ can be regarded as a synthesized artifact-affected image whose artifacts are from $x^a$ while whose content is from $y$. Thus, by reapplying $E_{I^a}^c$ and $G_I$, $y$ should be recovered.
	\begin{equation}
		G_{I}(E_{I^a}^c(\hat{y}^{a}))=\widetilde{y}\in I
	\end{equation}
	
	Based on these encoders and decoders, ADN uses four loss functions to optimize the output through artifact disentanglement; namely, an adversarial loss $L_{adv}$, an artifact consistency loss $L_{art}$, a reconstruction loss $L_{rec}$ and a self-reduction loss $L_{self}$. Then, the overall objective function is formulated as the weighted sum:
	\begin{equation}
		L = L_{adv}+\lambda_{rec}L_{rec}+\lambda_{art}L_{art}+\lambda_{self}L_{self}
	\end{equation}
	where the $\lambda$s are the hyperparameters controlling the relative importance of each loss. The loss functions are explained as follows.
	
	\subsubsection{Adversarial Loss}
	By manipulating the artifact code in the latent space, ADN outputs $\hat{x}$ and $\hat{y}^a$ where the former removes artifacts from $x^a$ and the latter adds artifacts to $y$. ADN adopts the strategy of adversarial learning by introducing two discriminators $D_{I^a}$ and $D_{I}$ to regularize the plausibility of $\hat{x}$ and $\hat{y}^a$ so that ADN can be trained without paired images. The adversarial loss can be written as
	\begin{eqnarray}
		&&L_{adv} = L_{adv}^{I}+L_{adv}^{I^a}\\
		&&where:\nonumber\\
		&&L_{adv}^{I} = {\mathbb{E}_{I}}[\log {D_I}(y)]+{\mathbb{E}_{I^a}}[1-\log {D_{I}}(\hat{x})]\nonumber\\
		&&L_{adv}^{I^a}= {\mathbb{E}_{I^a}}[\log {D_{I^a}}(x^a)]+{\mathbb{E}_{I,{I^a}}}[1-\log {D_{I^{a}}}(\hat{y}^{a})]\nonumber
	\end{eqnarray}
	
	\subsubsection{Reconstruction Loss}
	In a perfect artifact disentanglement process, encoding and decoding should not lose information nor introduce artifacts. For artifact reduction, after encoding and decoding by $E_{I^a}^c$ and $G_I$, the content information should be obtained. For artifact synthesis, using $E_{I^a}^a$, $E_I$ and $G_{I^a}$, an artifact-affected image should be generated. ADN utilizes two forms of reconstruction to encourage the encoders and decoders to preserve information. Specifically, it uses \{$E_{I^a}$, $G_{I^a}$ \} and \{$E_I$, $G_I$\} as autoencoders as follows:
	\begin{equation}
		L_{rec} = {\mathbb{E}_{I,{I^a}}}[\parallel \hat{x}^{a}-x^{a}  {\parallel _{{\rm{ 1}}}}+\parallel \hat{y}-y  {\parallel _{{\rm{ 1}}}} ]
	\end{equation}
	where the $L_1$ loss is used, instead of the $L_2$ loss, to encourage sharper outputs.
	\subsubsection{Artifact Consistency Loss}
	Since $\hat{y}^a$ is obtained by contaminating $y$ with the artifact $a$ from $x^a$ and $x$ is a clear image disentangled from $x^a$, $x^a$ and $y^a$
	contain the same artifact. In other words, we can obtain the same artifact from either $x^a-\hat{x}$ or $\hat{y}^a-y$. Thus, the artifact consistency loss can be introduced as follows:
	\begin{equation}
		L_{art} = {\mathbb{E}_{I,{I^a}}}[\parallel (x^{a}-\hat{x})-(\hat{y}^{a}-y)  {\parallel _{{\rm{ 1}}}} ]
	\end{equation}
	\subsubsection{Self-reduction Loss}
	If we add artifacts to $y$, which creates $\hat{y}^a$, and then remove the artifacts from $\hat{y}^a$, which results in $\widetilde{y}$, where $y$ and $\widetilde{y}$ should be close to each other. Based on this consideration, the following self-reduction loss is defined:
	\begin{equation}
		L_{self} = {\mathbb{E}_{I,{I^a}}}[\parallel \widetilde{y}-y  {\parallel _{{\rm{ 1}}}} ]
	\end{equation}
	
	\section{Proposed QS-ADN}
	Here we use ADN as a framework and modify it for quasi-supervised learning, which is referred to as QS-ADN.
	The proposed quasi-supervised learning model for LDCT image denoising includes two main parts: patch matching and network construction. Let us describe them in the following subsections.
	
	\subsection{Patch Matching}
	We use unpaired data of LDCT and NDCT images. Since image patches carry local information, we utilize them to determine the local similarity between patients. The target is to find the pairs of best matched patches and corresponding matching degrees as prior information to train our LDCT image denoising network. Because any patch from an LDCT image must be matched to NDCT image patches to find the best matching pairs, a high computational cost is needed. To address this problem, here we design an efficient method for a satisfactory performance. We first match slices from LDCT and NDCT scans and then match patches from the matched slices. The workflow is described as follows:
	\begin{enumerate}
		\item For a given slice in an LDCT image set, the best-matched slice in an NDCT image set is identified using a similarity measure.
		\item For a given patch in an LDCT slice, the best-matched patch is identified in the corresponding matched NDCT slice using the similarity measure.
		\item The corresponding LDCT and NDCT patch pairs are obtained, and the similarity degrees are computed.
	\end{enumerate}
	

		In the matching process, we can use the normalized mutual information (NMI), the Pearson correlation coefficient, the radial basis function (RBF) and other functions to measure the similarity between two image patches. 
		
		The standard mutual information is defined as follows:
		
		\begin{equation}
			I(X,Y) = \sum\limits_{x } {\sum\limits_{{ y} } p } (x,y)log\frac{{p(x,y)}}{{p(x)p(y)}}
		\end{equation}
		where $X$ and $Y$ form an image pair, $p(x)$ and $p(y)$ denote the distributions of $X$ and $Y$ respectively, and $p(x,y)$ denotes the joint distribution of $X$ and $Y$. Then, NMI is formulated as
		
		\begin{equation}
			NMI(X,Y) = \frac{{2I(X,Y)}}{{H(X) + H(Y)}}
		\end{equation}
		where $H(X) =  - \sum\limits_i p ({x_i})\log p({x_i})$.

		The Pearson correlation coefficient is expressed as	
		\begin{equation}	
			\rho (x,y) = \frac{{{\mathop{\rm{\emph c\emph o \emph v}}} (x,y)}}{{{\sigma _x}{\sigma _y}}} 
		\end{equation}	
		where $cov(x,y)$ is the covariance of $x$ and $y$, and $\sigma _x$ and $\sigma _y$ are the standard deviations of $x$ and $y$ respectively.
		
		The RBF is defined as follows:	
		\begin{equation}	
			RBF(x,y) = {e^{ - \frac{{\left\| {\left. {x - y} \right\|} \right._2^2}}{{2{\sigma ^2}}}}}
		\end{equation}
		where $\sigma>0$ is a hyperparameter.
		
		The ultimate goal in this process is to find the pairs of most similar patches. The slice matching step sacrifices precision to improve efficiency. When the computational resources are sufficient, this step can be removed. Additionally, a patch under a given imaging condition could be matched to one or more patches under different imaging conditions, which can help improve the denoising ability of the neural network.
		
		\subsection{Network Construction}
		Image synthesis is an important process for ADN, which uses any unpaired artifact-free and artifact-affected images ($x^a$ and $y$) to generate a new image. Certainly, if $x^a$ and $y$ are paired, the performance of ADN will become better than that of the generic ADN with unpaired images; however, much extra work is required to acquire paired data, as discussed above. We recall our previous work on obtaining the pairs of best matched patches from unpaired data and recording the matching degrees. These matching degrees will be used here to improve ADN, which leads to our proposed QS-ADN. Specifically, we add the matching degree information $(x^a,y)$ to weight the corresponding losses.
		By observation, we identify that three losses ($L_{rec}$, $L_{art}$ and $L_{self}$) are simultaneously related to both $x^a$ and $y$, which should be weighted. $L_{adv}$ indicates the difference between two sets sampled from $I$ and $I^a$, which should not be weighted.
		Therefore, the loss function of QS-ADN can be expressed as
		
		\begin{equation}
			L =  L_{adv}+w(I,I^a)\otimes(\lambda_{rec}L_{rec}+\lambda_{art}L_{art}+\lambda_{self}L_{self})\label{eq:total_loss}
		\end{equation}
		where $w(I,I^a)$ denotes the weights corresponding to the similarity of two image patches from $I$ and $I^a$ respectively, and the operator $\otimes$ assigns these weights to the corresponding components in the loss function, which has been interpreted in Eq. \eqref{eq:qsl}. All the involved encoders, decoders and discriminators have the same structures as that used in the classic ADN.
		
		Traditionally, the patch pairs used in deep learning have only two statuses: paired and unpaired. In contrast to this binary classification, quasi-supervised learning considers the degree of patch matching. This is more in line with a fuzzy reasoning process. As a special case, such pairs of patches form a supervised training dataset when only the pairs with a matching probability of one are included. Therefore, our method is compatible with supervised and semi-supervised learning modes.
		
		\section{Experiments}	
		
		To verify the denoising effect of QS-ADN, we used the Mayo chest CT image dataset from The Cancer Imaging Archive (TCIA) \cite{data}, which is an open and vendor-neutral CT patient database created by the Mayo Clinic. The dataset contains paired LDCT and NDCT images, where quarter-dose and full-dose filtered backprojection images
		are provided with the corresponding projection datasets.
		We selected ten patients as the training dataset and two other patients as the testing dataset. In the training set, the LDCT images of five patients are taken as input, and the NDCT images of the remaining five patients are taken as the learning target. The size of each image is 512$\times$512.

		The dataset contains paired LDCT and NDCT images and, thus, enables qualitative and quantitative evaluation. We use the peak signal-to-noise ratio (PSNR) \cite{51} to measure the denoising performance by calculating the overall difference between the denoised LDCT image and the original NDCT image. A higher value indicates better image quality. The PSNR is expressed as
		\begin{equation}
			PSNR = 10{\log _{10}}\left( {\frac{{{\rm{ma}}{{\rm{x}}^2}}}{{\frac{1}{n}\sum\nolimits_{i = 1}^n {{{\left( {{x_i} - {y_i}} \right)}^2}} }}} \right)
		\end{equation}
		where $x_i$ and $y_i$ are pixel values of LDCT and NDCT images respectively, $n$ is the number of pixels in an image, and $max$ represents the maximum image pixel value.
		
		We also use the structural similarity (SSIM) \cite{52} to evaluate the denoised image quality. The value is in the range of $[-1,1]$, and the higher the value is, the closer the image features are to that in the NDCT image.
		
		\begin{equation}
			SSIM(x,y)=\frac{\left(2\mu_x\mu_y+c_1\right)\left(\sigma_{xy}+c_2\right)}
			{\left(\mu_x^2+\mu_y^2+c_1\right)\left(\sigma_x^2+\sigma_y^2+c_2\right)}
		\end{equation}
		where $\mu_x$ and $\mu_y$ are the averages of $x$ and $y$, $\sigma_x^2$ and $\sigma_y^2$ are the standard deviations of $x$ and $y$ respectively, $\sigma_{xy}$ is the covariance of $x$ and $y$, and $c_1$ and $c_2$ are two offset constants to stabilize the division operation.
		
		We use NMI to measure the similarity of two patches when selecting the matched pairs. Since the range of NMI is $[0,1]$, we directly used NMI values as the weights ($w(x,y)$) for network construction. 
		
		The network was implemented in Python 3.7 with PyTorch 1.2.0 on a computer equipped with an NVIDIA Quadro P2200 GPU. We used the Adam optimizer to optimize the loss function, and the learning rate was set to 0.0001 with $\beta_1=0.5$ and $\beta_2=0.99$.
		
		\subsection{NMI Distribution}
\begin{figure*}
			\centering
			\subfigure[]
			{\includegraphics[width=0.45\textwidth]{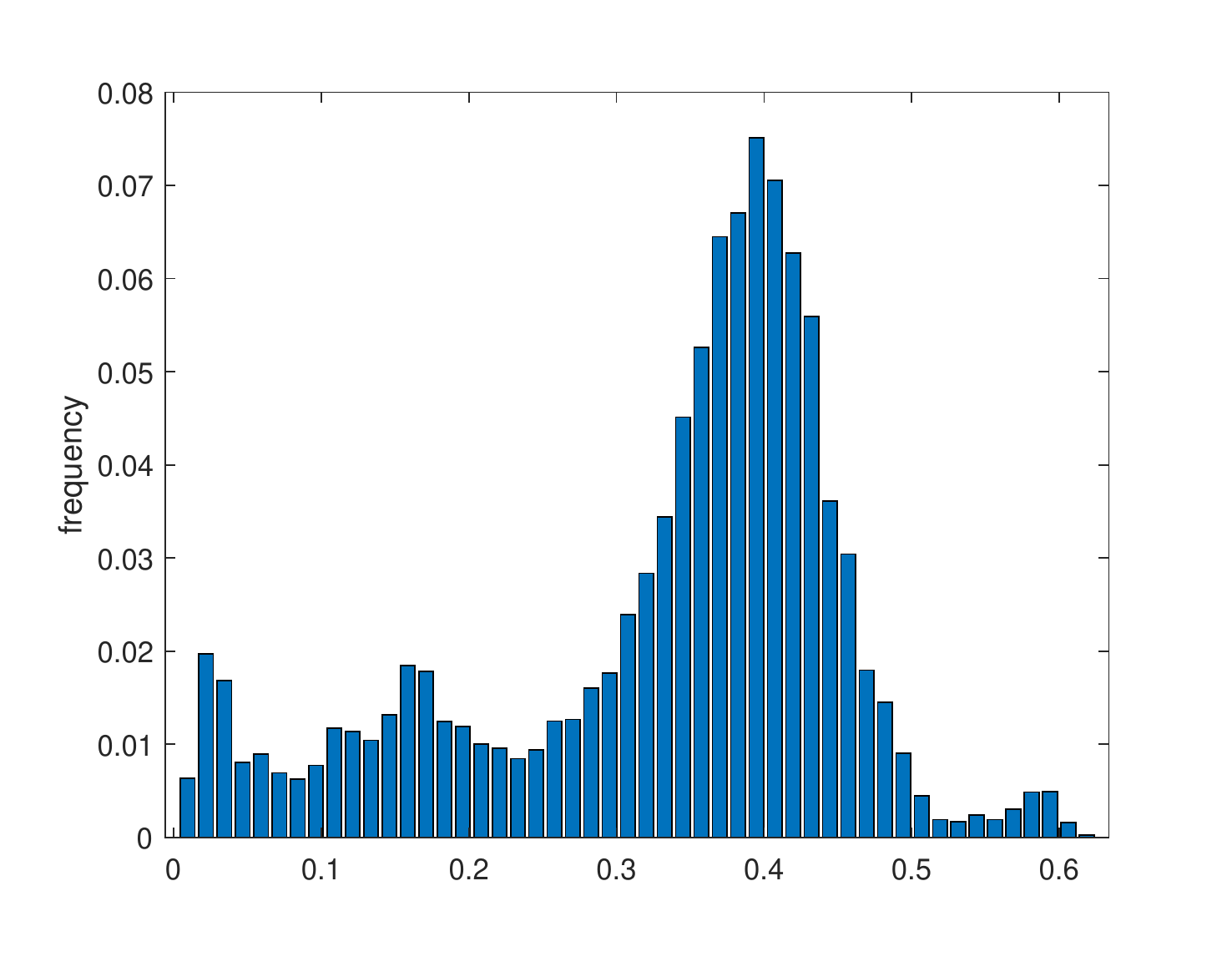}}
			\subfigure[]
			{\includegraphics[width=0.45\textwidth]{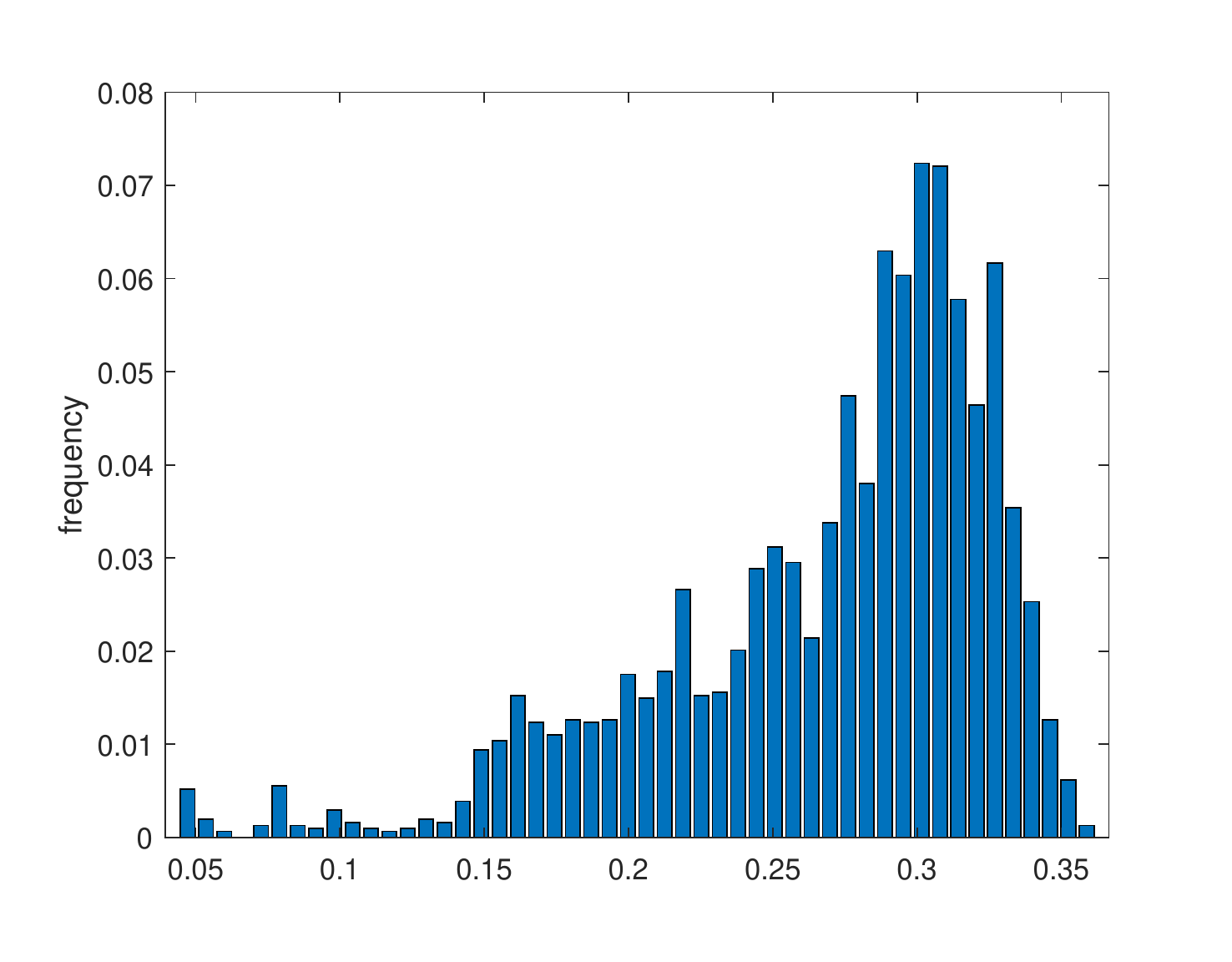}}
			\caption{Distributions of NMI values for (a) truly paired data and (b) computationally paired data.}
			\label{fig13}
		\end{figure*}

		We first compared the NMI distribution of the truly and manually paired data, which is shown in
		Fig. \ref{fig13}.
		Most of the truly paired data have NMI values of approximately 0.4, and the manually paired data have values of approximately 0.31. The former certainly and clearly have higher values than the latter. However, the two distributions overlap in the wide range of $[0.1,~0.35]$. Since low matching degrees are not significant for QS-ADN, we only saved the pairs with matching degrees above 0.1 for training. In fact, the NMI values of the truly paired data are not as high as expected, and some values are very low, approaching 0.01. Such low NMI values for the paired data provide an opportunity for our method to catch up under guidance of the matching degree information.
		The matching degrees of the manually paired data are actually related to the scale of the data. With a larger dataset, the matching degrees usually become higher up to the case of truly paired data. 
		In fact, according to the following experiments, such similarity degrees are sufficient for satisfactory denoising performance.

		\subsection{Hyperparameter Selection}
		In consideration of the computer hardware used in this study, we set the training batch size to 2. As suggested by \cite{Ref42}, we set the hyperparameters as follows: $\lambda_{rec}=\lambda_{art}=\lambda_{self}=\lambda$. Then, several key factors of the proposed QS-ADN were studied for LDCT image denoising, including the patch size, number of training epochs, and weighting hyperparameters ($\lambda$).
		
		\subsubsection{Patch Size}
\begin{figure*}
			\centering
			\subfigure[]
			{\includegraphics[width=0.45\textwidth]{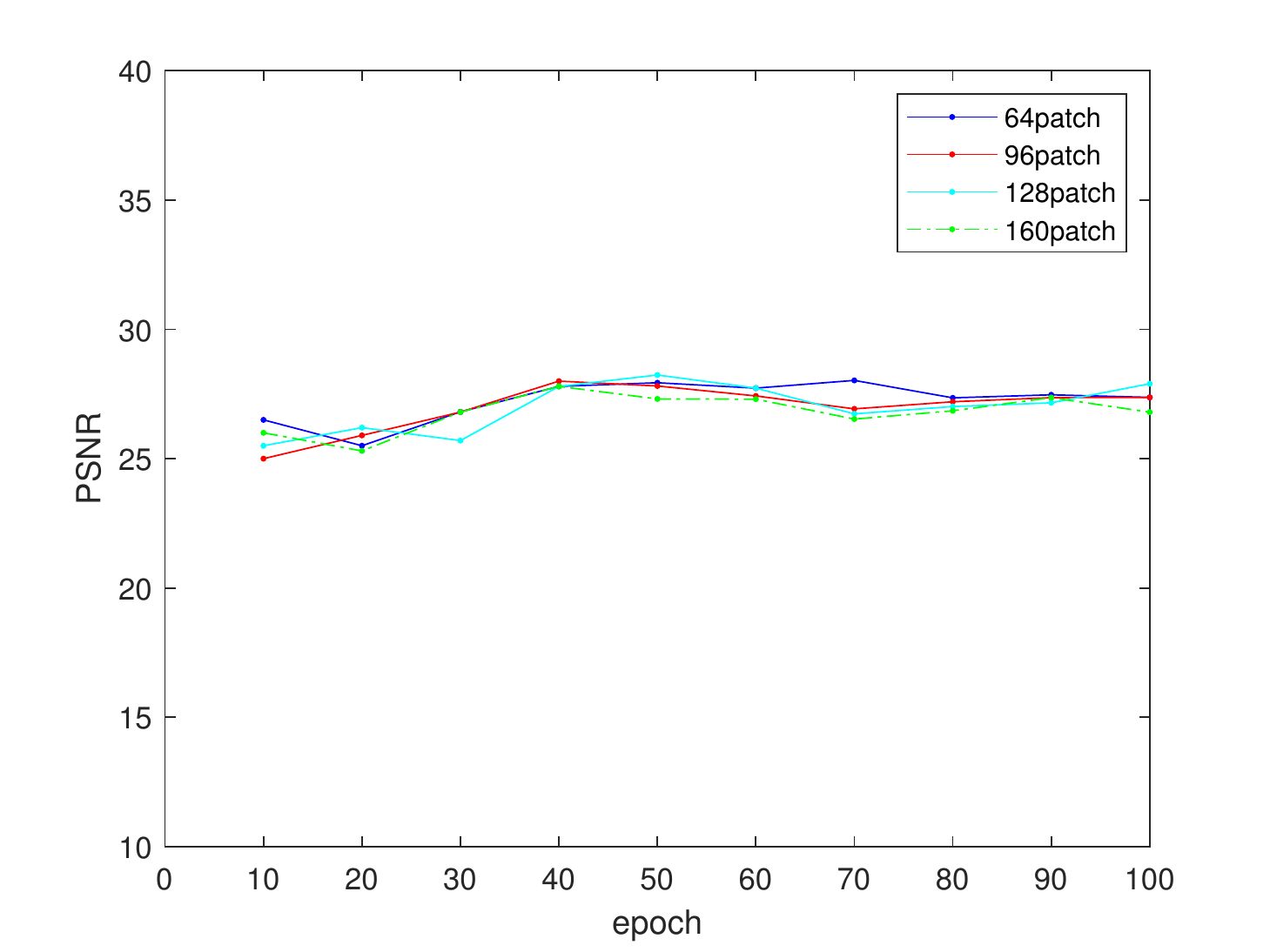}}
			\subfigure[]
			{\includegraphics[width=0.45\textwidth]{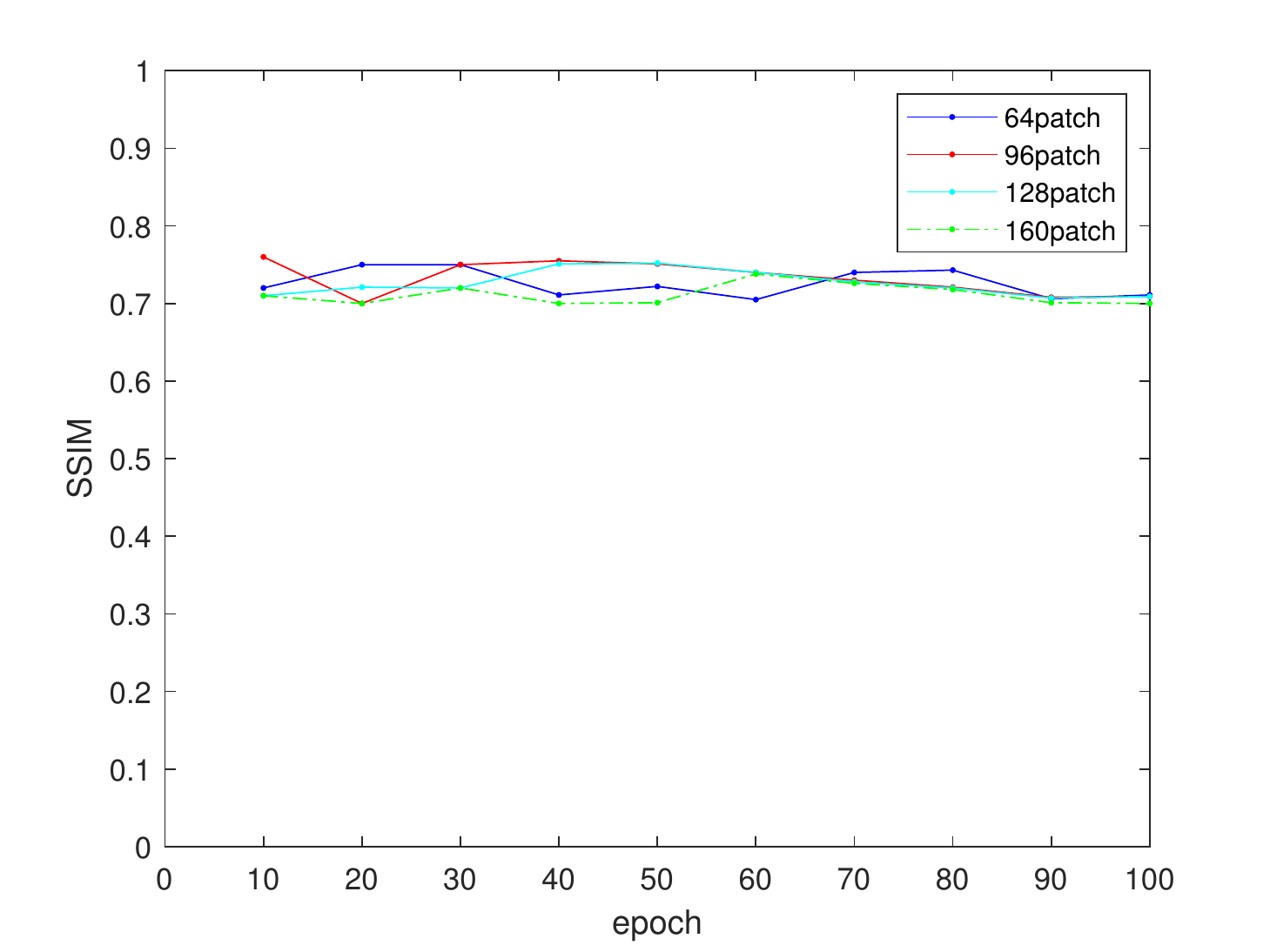}}
			\caption{Evaluative metrics versus patch size with different numbers of epochs: (a) PSNR and (b) SSIM. }
			\label{fig3}
		\end{figure*}

		Quasi-supervised learning is based on patch pairs. A patch is an important local information carrier, and its size may affect the capability of extracting contextual information.  We performed experiments to study the effect of the patch size on the denoising results as shown in Fig. \ref{fig3}, where $\lambda=20$ was used. We gradually increased the patch size from 64 to 160 with different numbers of epochs. It was found that the size of the patch had no major effect on the denoising results for the proposed network. Based on this observation, given the computer hardware for reasonable training time we fixed the patch size to 64$\times$64 for the subsequent experiments.
		
		\subsubsection{Number of Epochs}
\begin{figure}
			\centering
			\includegraphics[width=0.5\textwidth]{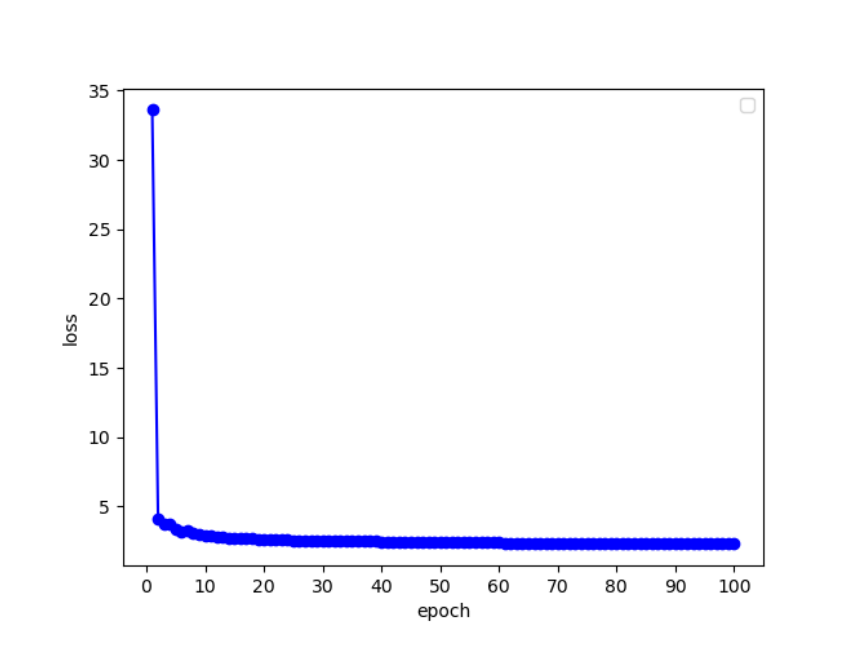}
			\caption{Loss curve of QS-ADN on the training set.}
			\label{fig4}
		\end{figure}

		Choosing a suitable number of training epochs is crucial for ensuring network convergence and saving training time.
		To obtain a satisfactory denoising result, we usually iterated the network training process sufficiently many times. However, when the denoising effect is similar, additional iterations is not needed to cut computational cost. The loss curve on the training set is shown in Fig. \ref{fig4}. We still set $\lambda=20$. The loss curve rapidly decreased in the first 10 epochs, especially in the first two, and became quite flat after 60 epochs. Considering the denoising effect and computational cost, we set the number of epochs to 70 in the subsequent experiments.

		\subsubsection{Value of $\lambda$}
\begin{figure*}
			\centering
			\subfigure[]
			{\includegraphics[width=0.45\textwidth]{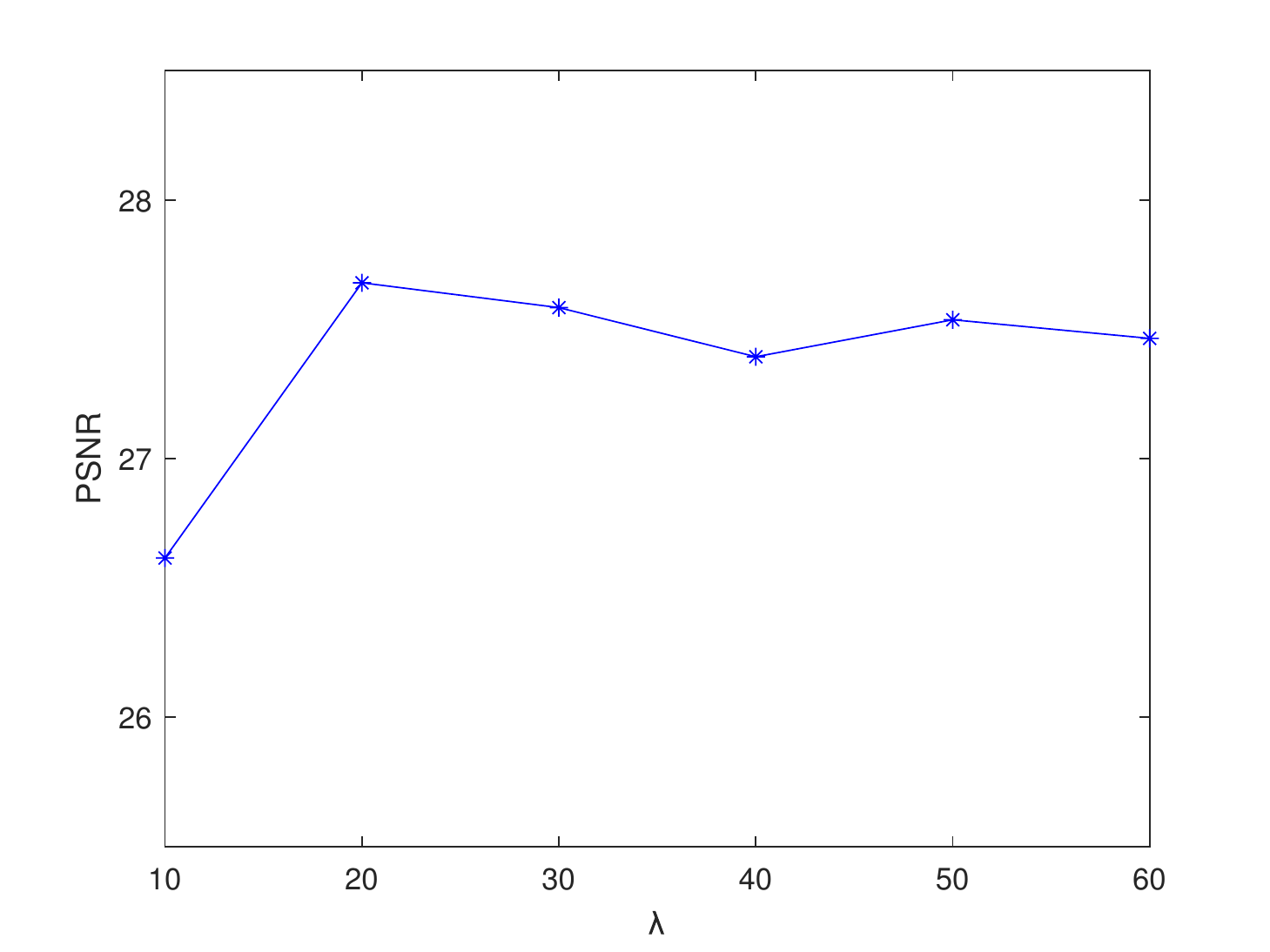}}
			\subfigure[]
			{\includegraphics[width=0.45\textwidth]{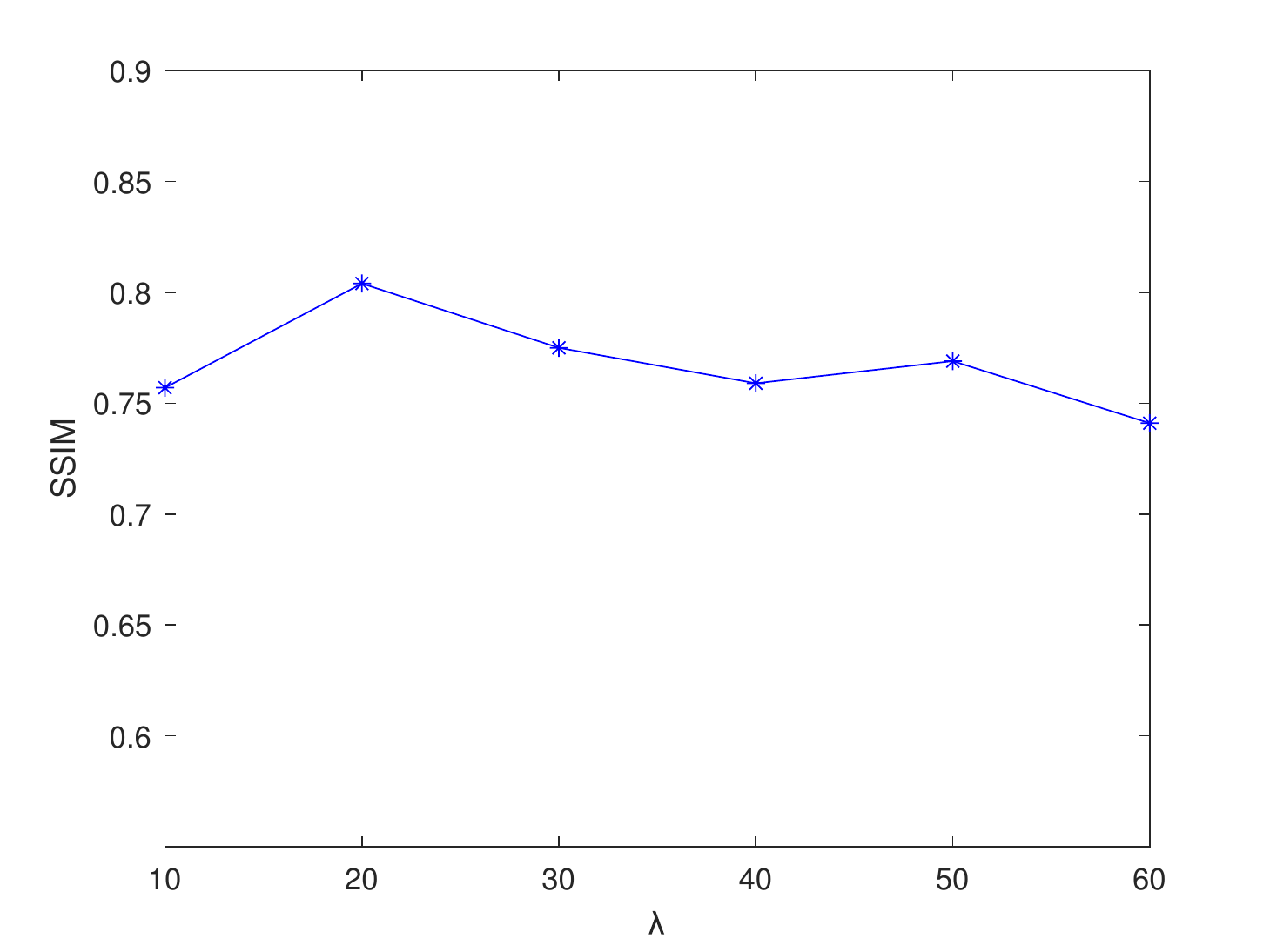}}
			\caption{Evaluative metrics versus $\lambda$: (a) PSNR and (b) SSIM.}
			\label{fig4.3}
		\end{figure*}

		It is important to select a proper value for the hyperparameter $\lambda$ in the total loss function, which indicates the importance of the quasi-supervision dependent components.
		We selected the optimal value from the range of $ \{10,20,30,40,50,60\}$. As shown in Fig. \ref{fig4.3}, by the evaluation with both PSNR and SSIM, $\lambda = 20$ was identified as the optimal value. In general, the PSNR and SSIM curves showed similar trends.
		
		\subsection{Noise Synthesis}	
\begin{figure*}
			\centering
			\subfigure[]
				{\includegraphics[width=0.49\textwidth]{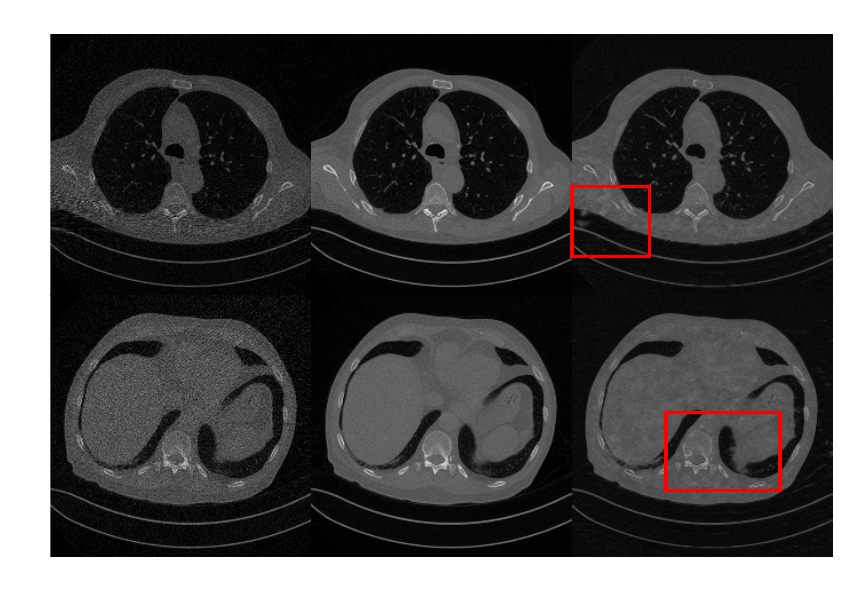}}
			\subfigure[]
						{\includegraphics[width=0.49\textwidth]{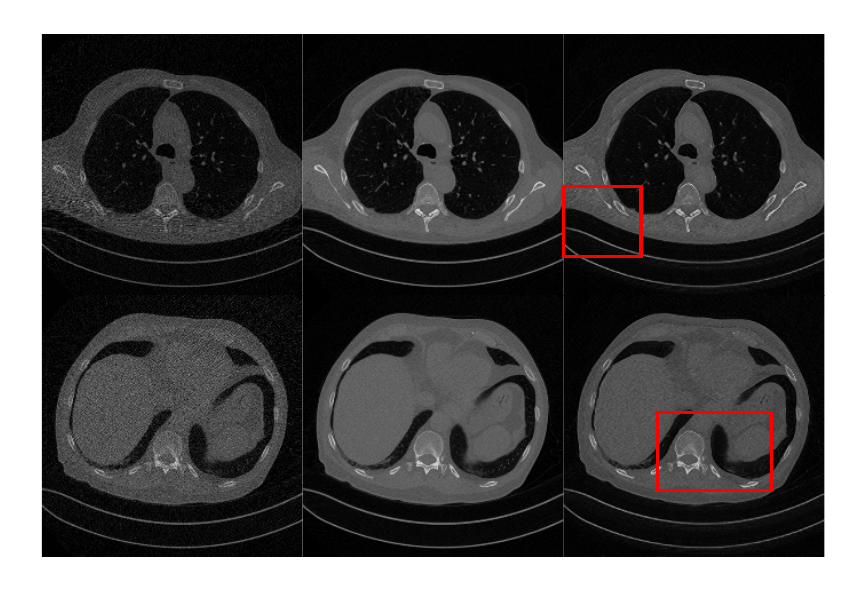}}
			\caption{Noisy image synthesis by (a) ADN and (b) QS-ADN. Left: LDCT images; middle: NDCT images; right: synthetic LDCT images obtained from NDCT images with the noise originating from unpaired LDCT images}
			\label{fig5}
		\end{figure*}

		In addition to noise reduction, ADN also supports noise synthesis. A good synthetic noisy image is very helpful for noise reduction. Fig. \ref{fig5} shows synthetic noisy images generated by ADN and QS-ADN. The synthetic noise introduced by ADN caused severe damage to anatomical structures in the NDCT images. In contrast, QS-ADN produced more realistic noise and more similar anatomical structures in reference to the actual noisy images. In particular, as marked in the figure, the anatomical structures were more seriously damaged by ADN than that by QS-ADN.
		The main reason is that ADN used randomly paired images to train the network, while QS-ADN used optimally matched images with their matching degrees to train the network more purposely.
		
		\subsection{Denoising Results}
		We compared six state-of-the-art methods with our proposed method: BM3D, median filtering with window size $3\times3$, Gaussian low-pass filtering with $\sigma=55$ in the frequency domain, DualGAN \cite{DualGAN}, CycleGAN \cite{Ref32} and Noise2Sim \cite{n2sim}. BM3D is a popular image-based denoising method and has been applied to LDCT image denoising. Gaussian low-pass and median filtering are both basic image denoising methods. DualGAN and CycleGAN are unsupervised deep learning methods based on the GAN framework in the field of image-to-image translation. Noise2Sim is a self-learning method for image denoising that leverages self-similarities of image patches and learns to map between the center pixels of similar patches for self-consistent image denoising, where we search similar patches in two-dimensional mode. 
		
\begin{figure}
			\centering
			\includegraphics[width=0.49\textwidth]{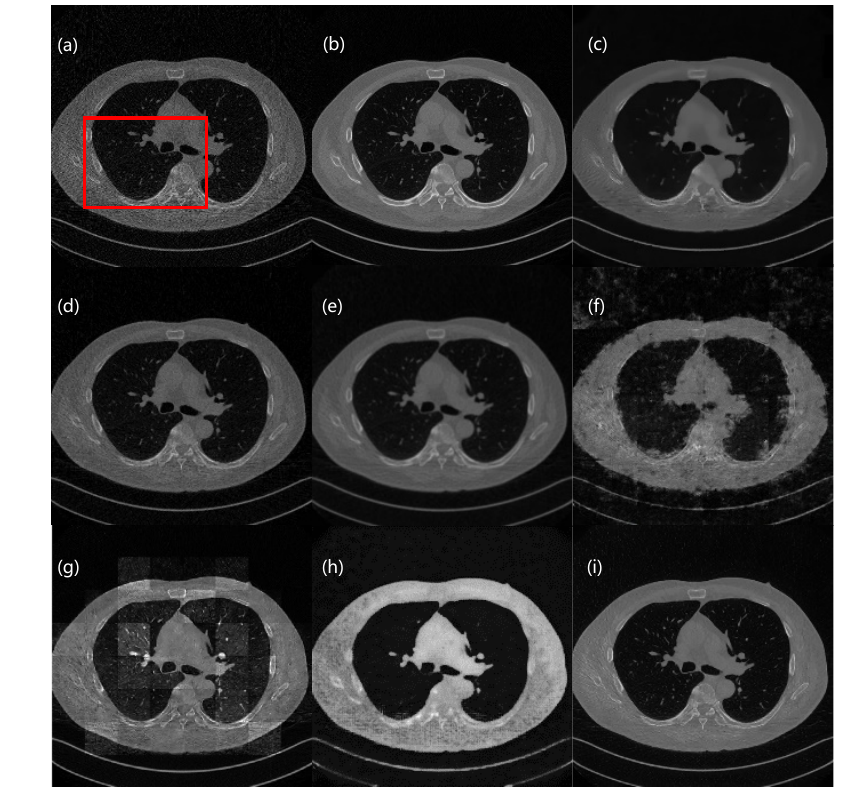}
			\caption{Denoising results: (a) LDCT, (b) NDCT, (c) BM3D, (d) median filtering, (e) Gaussian low-pass filtering, (f) DualGAN, (g) CycleGAN, (h) Noise2Sim, and (i) QS-ADN.}
			\label{fig7}
		\end{figure}

\begin{figure}
			\centering
			\includegraphics[width=0.49\textwidth]{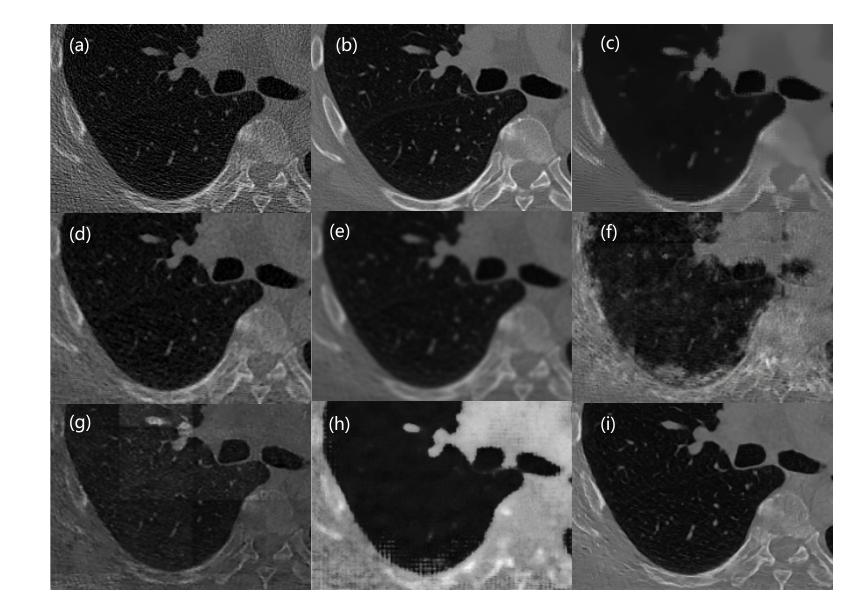}
			\caption{Enlarged ROI marked by the red box in Fig. \ref{fig7}: (a) LDCT, (b) NDCT, (c) BM3D, (d) median filtering, (e) Gaussian low-pass filtering, (f) DualGAN, (g) CycleGAN, (h) Noise2Sim, and (i) QS-ADN.}
			\label{fig8}
		\end{figure}

		To evaluate the denoising performance, we selected a representative slice from the chest area, as shown in Fig. \ref{fig7}. Evidently, the denoising results of BM3D and Gaussian low-pass filtering are too smooth, and many details were lost. Median filtering did not produce satisfactory results and still generated considerable noise. DualGAN, CycleGAN and Noise2Sim removed considerable noise but also lost details. The results by DualGAN were a bit rough, being substantially different from the real NDCT image. CycleGAN gave a better result than DualGAN, which, however, is still distorted seriously compared to that of NDCT. Noise2Sim provided over-smoothed results. In contrast to these results, QS-ADN was able to retain most details while suppressing noise effectively. It has better clarity than other denoising results, and seems very similar to the true NDCT image. An enlarged image in the marked region of interest (ROI) is displayed in Fig. \ref{fig8}, from which we can more clearly observe the detail of the images. By comparsion, we further verifies the conclusion drawn from the above
		that the proposed QS-ADN indeed provides the best reconstructed image.


			For a systematic analysis, we quantitatively compared the denoising performance of these methods on the whole testing set in Tab. \ref{tab:1}.
			Among them, amazingly CycleGAN obtained the lowest PSNR and SSIM values, and its performance is even worse than qualitatively illustrated in Figs. \ref{fig7} and \ref{fig8}.
			DualGAN and Noise2Sim obtained similar quality metric values. The traditional filtering methods outperformed DualGAN and Noise2Sim. The proposed QS-ADN offerred the best quality metric values regradless of PSNR or SSIM, showing its ability to learn the intrinsic mapping from LDCT to NDCT images.  Nevertheless, we would like to note that Noise2Sim can denoise any noisy images individually without using paired or unpaired datasets at all, which only requires LDCT images.
			
			\begin{table}
				\centering
				\caption{Quantitative evaluation of competing denoising methods on the whole testing set, where the best results are in bold.}
				\label{tab:1}       
				\begin{tabular}{llllll}
					\hline\noalign{\smallskip}
					Method&PSNR & SSIM \\
					\noalign{\smallskip}\hline\noalign{\smallskip}
					BM3D & 24.693 & 0.800\\
					Median filtering & 24.861 & 0.784 \\
					Gaussian low-pass filtering & 23.059 & 0.735 \\
					DualGAN &22.101 &0.453 \\
					CycleGAN &18.022 & 0.309 \\
					Noise2Sim & 22.887 & 0.748\\
					QS-ADN & \bf{27.680} & \bf {0.804} \\
					\noalign{\smallskip}\hline
				\end{tabular}
			\end{table}

			%

			\subsection{Ablation Studies}
\begin{figure}
					\centering
					\includegraphics[width=0.49\textwidth]{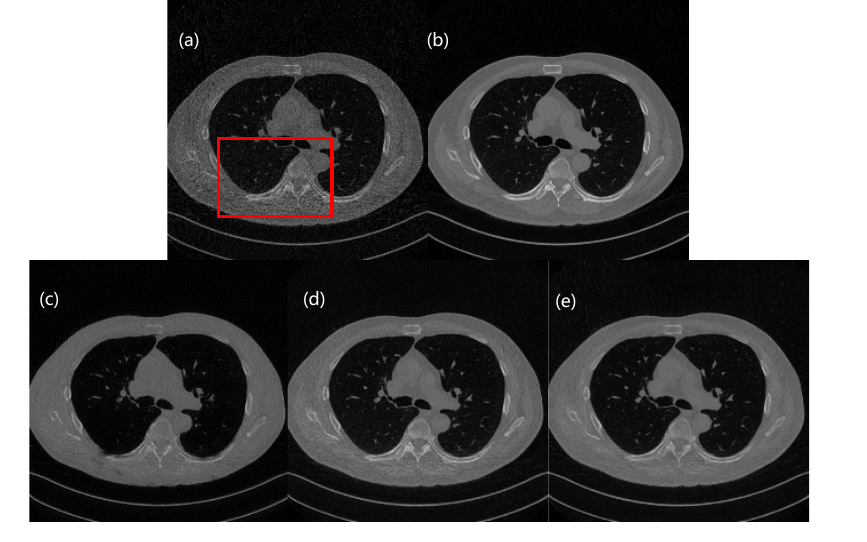}
					\caption{Qualitative comparison: (a) LDCT, (b) NDCT, (c) ADN, (d) ADN+patch matching, and (e) QS-ADN.}
					\label{fig11}       
				\end{figure}

\begin{figure}
					\centering
					\includegraphics[width=0.49\textwidth]{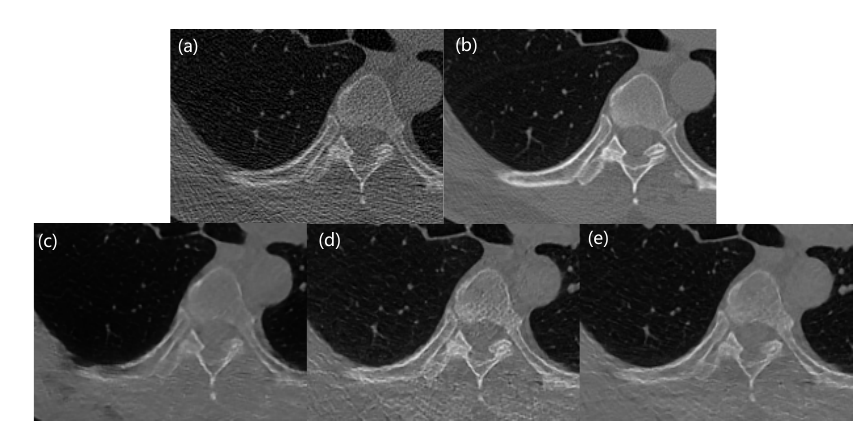}
					\caption{Enlarged ROI images marked by the red box in Fig. \ref{fig11}: (a) LDCT, (b) NDCT, (c) ADN, (d) ADN+patch matching, and (e) QS-ADN.}
					\label{fig12}       
				\end{figure}

			The proposed QS-ADN consists of the following three key parts:
			\begin{equation}
				QS-ADN=ADN + patch~matching+ weighting\nonumber
			\end{equation}
			We conducted an ablation study to assess the effectiveness of each part.
			Fig. \ref{fig11} shows the denoised images obtained by ADN, ADN$+$patch matching and QS-ADN, and Fig. \ref{fig12} enlarged the ROI. By comparing the LDCT and denoised images, it can be seen that there is substantial noise contaminating the LDCT image; for example, anatomical details in the marked ROI are almost invisible. ADN indeed provided a decent denoising result; however, oversmoothing was substantial such that the small tissue structures were lost. The combination of ADN and patch matching
			reduced the over-smoothing effect compared with ADN, providing a more detailed appearance that thanks to patch matching.
			Remarkably, QS-ADN further improved the performance such that small tissue features can be seen clearly while suppressing the image noise much better than the other two options. This demonstrates the importance of our proposed weighting scheme.
			


				As before, we further performed the statistical analysis on the whole testing set for a systematic comparison. Tab. \ref{tab:2} indicates that QS-ADN outperformed ADN$+$patch matching, and the latter outperformed the generic ADN, being consistent with our previous observations.
				\begin{table}
					\centering
					\caption{Quantitative analysis results of quasi-supervised learning with different combinations of the key modules, where the best results are in bold.}
					\label{tab:2}       
					\begin{tabular}{lll}
						\hline\noalign{\smallskip}
						
						Method & PSNR & SSIM  \\
						\noalign{\smallskip}\hline\noalign{\smallskip}
						ADN & 26.639& 0.738 \\
						ADN+patch matching&  27.538 & 0.755\\
						QS-ADN& \bf{27.680} & \bf{0.804} \\
						\noalign{\smallskip}\hline
					\end{tabular}
				\end{table}

				\section{Discussions and Conclusion}

				Clearly, a larger dataset better reflects the diversity of tissue structure and covers more similar structures. Limited by our available dataset and computational resources, this paper only used a moderately-sized dataset. In the future, we will use a larger number of unpaired CT images to improve the denoising performance further in the quasi-supervised learning mode.
				
				In conclusion, we have generalized ADN into a quasi-supervised version, which takes advantage of massive unpaired datasets that contain similar local structures. Our proposed method avoids the difficulty of obtaining supervised/labeled CT data, increasing neither radiation dose nor imaging cost. This is a new and cost-effective approach for LDCT image denoising.
				
				
				\balance
				\bibliographystyle{IEEEtran}
				
				\bibliography{reference}
				
			\end{document}